\documentclass [11pt]{article}
\input{psfig.sty}
\usepackage{graphicx}
\usepackage{hyperref}

\begin{document}

\title{Multi Expression Programming - an in-depth description}
\label{mep_chapter}

\author{Mihai Oltean\\
\\
Department of Computer Science, Babe\c s-Bolyai University\\
Kogalniceanu 1, Cluj-Napoca, 400084, Romania\\
\\
mihai.oltean@gmail.com\\
\url{https://mepx.org}
}

\date{June 4, 2006}

\maketitle


Multi Expression Programming (MEP) \cite{oltean_complex, oltean_mep_eea,oltean_improving,oltean_circuits_nasa} is a Genetic Programming variant that uses a linear representation of chromosomes. MEP individuals are strings of genes encoding complex computer programs.

When MEP individuals encode expressions, their representation is similar to 
the way in which compilers translate $C$ or \textit{Pascal} expressions into machine code \cite{aho1}. 

A unique MEP feature is the ability of storing multiple solutions of a 
problem in a single chromosome. Usually, the best solution is chosen for 
fitness assignment. When solving symbolic regression or classification 
problems (or any other problems for which the training set is known before 
the problem is solved) MEP has the same complexity as other techniques 
storing a single solution in a chromosome (such as GP, CGP, GEP or GE).

Evaluation of the expressions encoded into a MEP individual can be performed 
by a single parsing of the chromosome.

Offspring obtained by crossover and mutation are 
always syntactically correct MEP individuals (computer programs). Thus, no 
extra processing for repairing newly obtained individuals is needed.

\section{Chapter structure}

The chapter is structured as follows:

MEP representation is described in section \ref{MEP_representation}. The way in which a MEP chromosome is initialized is described in section \ref{MEP_initialization}.

Fitness assignment process is described in section \ref{MEP_fitness}.

Genetic operators used in conjunction with MEP are introduced in section \ref{MEP_Search_operators}. 

MEP representation is compared to other GP representations in section \ref{MEPvsOthers}.

The way in which MEP handles exceptions is described in section \ref{MEP_exceptions}.

The complexity of the fitness assignment process is computed in section \ref{MEP_complexity}.

MEP algorithm is given in section \ref{MEP_algorithm}.

Automatically Defined Function in MEP are introduced in section \ref{mep_adfs}.

Tips for a good implementation of the MEP technique are suggested in section \ref{MEP_problems}.

Further research directions and open problems are indicated in section \ref{MEP_problems}.

\section{Representation}\label{MEP_representation}

MEP genes are (represented by) substrings of a variable length. The number 
of genes per chromosome is constant. This number defines the length of the 
chromosome. Each gene encodes a terminal or a function symbol. A gene that 
encodes a function includes pointers towards the function arguments. 
Function arguments always have indices of lower values than the position of 
the function itself in the chromosome.

MEP representation ensures that no cycle arises while the 
chromosome is decoded (phenotypically transcripted). According to the 
MEP representation scheme, the first symbol of the chromosome must be a 
terminal symbol. In this way, only syntactically correct programs (MEP 
individuals) are obtained.\\

\textbf{Example}\\

Consider a representation where the numbers on the left positions stand for 
gene labels. Labels do not belong to the chromosome, as they are provided 
only for explanation purposes.

For this example we use the set of functions:\\

$F$ = {\{}+, *{\}},\\

\noindent
and the set of terminals\\

$T$ = {\{}$a$, $b$, $c$, $d${\}}.\\

An example of chromosome $C$ using the sets $F$ and $T$ is given below:\\

1: $a$

2: $b$

3: + 1, 2

4: $c$

5: $d$

6: + 4, 5

7: * 3, 5

8: + 2, 6\\

The maximum number of symbols in MEP chromosome is given by the formula:\\

\textit{Number{\_}of{\_}Symbols} = ($n + $1) * (\textit{Number{\_}of{\_}Genes} -- 1) + 1, \\

\noindent
where $n$ is the number of arguments of the function with the greatest number 
of arguments. 

The maximum number of effective symbols is achieved when each gene 
(excepting the first one) encodes a function symbol with the highest number 
of arguments. The minimum number of effective symbols is equal to the number 
of genes and it is achieved when all genes encode terminal symbols only.\\

\section{Initialization}
\label{MEP_initialization}

There are some restrictions for generating a valid MEP chromosome:

\begin{itemize}

\item{First gene of the chromosome must contain a terminal. If we have a function in the first position we also need some pointer to some positions with lower index. But, there are no other genes above the first gene.}
\item{For all other genes which encodes functions we have to generate pointers toward function arguments. All these pointers must indicate toward genes which have a lower index than the current gene.}

\end{itemize}

\section{Fitness assignment}
\label{MEP_fitness}

Now we are ready to describe how MEP individuals are translated into 
computer programs. This translation represents the phenotypic transcription 
of the MEP chromosomes.

Phenotypic translation is obtained by parsing the chromosome top-down. A 
terminal symbol specifies a simple expression. A function symbol specifies a 
complex expression obtained by connecting the operands specified by the 
argument positions with the current function symbol.

For instance, genes 1, 2, 4 and 5 in the previous example encode simple 
expressions formed by a single terminal symbol. These expressions are:\\

$E_{1} = a$,

$E_{2} = b$,

$E_{4} = c$,

$E_{5} = d$,\\

Gene 3 indicates the operation + on the operands located at positions 1 and 
2 of the chromosome. Therefore gene 3 encodes the expression:\\

$E_{3}=a+b$.\\

Gene 6 indicates the operation + on the operands located at positions 4 and 
5. Therefore gene 7 encodes the expression:\\

$E_{6}=c+d$.\\

Gene 7 indicates the operation * on the operands located at position 3 and 
5. Therefore this gene encodes the expression:\\

$E_{7}=(a+b)*d$.\\

Gene 8 indicates the operation + on the operands located at position 2 and 
6. Therefore this gene encodes the expression:\\

$E_{8}=b*(c+d)$.\\

There is neither practical nor theoretical evidence that one of these 
expressions is better than the others. Moreover, Wolpert and McReady \cite{wolpert1,wolpert2} 
proved that we cannot use the search algorithm's behavior so far for a 
particular test function to predict its future behavior on that function. 
This is why each MEP chromosome is allowed to encode a number of expressions 
equal to the chromosome length (number of genes). The chromosome described 
above encodes the following expressions:\\

$E_{1}=a$,

$E_{2}=b$,

$E_{3}=a+b$,

$E_{4}=c$,

$E_{5}=d$,

$E_{6}=c+d$,

$E_{7} = (a+b) * d$,

$E_{8}=b*(c+d)$.\\

The value of these expressions may be computed by reading the chromosome top 
down. Partial results are computed by Dynamic Programming \cite{bellman1} and are stored 
in a conventional manner.

\begin{figure}[htbp]
\centerline{\includegraphics[width=3.82in,height=5.26in]{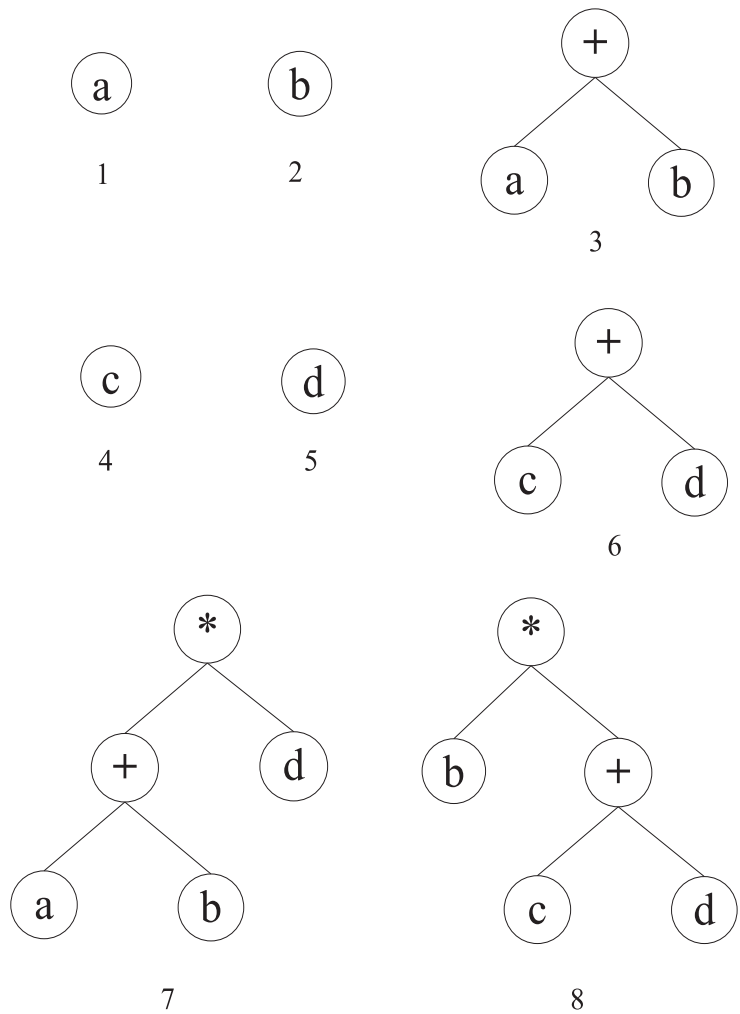}}
\label{mep:fig1}
\caption{Expressions encoded by a MEP chromosome represented as trees}
\end{figure}

Due to its multi expression representation, each MEP chromosome may be 
viewed as a forest of trees rather than as a single tree, which is the case 
of Genetic Programming. Figure \ref{mep:fig1} shows the forest of expressions encoded by the previously presented MEP chromosome. 

As MEP chromosome encodes more than one problem solution, it is interesting 
to see how the fitness is assigned. 

The chromosome fitness is usually defined as the fitness of the best 
expression encoded by that chromosome.

For instance, if we want to solve symbolic regression problems, the fitness 
of each sub-expression $E_{i}$ may be computed using the formula:

\begin{equation}
\label{eq1}
f(E_i ) = \sum\limits_{k = 1}^n {\left| {o_{k,i} - w_k } \right|} ,
\end{equation}

where $o_{k,i}$ is the result obtained by the expression $E_{i}$ for the 
fitness case $k$ and $w_{k}$ is the targeted result for the fitness case $k$. In 
this case the fitness needs to be minimized.

The fitness of an individual is set to be equal to the lowest fitness of the 
expressions encoded in the chromosome:

\begin{equation}
\label{eq2}
f(C) = \mathop {\min }\limits_i f(E_i).
\end{equation}

When we have to deal with other problems, we compute the fitness of each 
sub-expression encoded in the MEP chromosome. Thus, the fitness of the 
entire individual is supplied by the fitness of the best expression encoded 
in that chromosome. An example on how to compute the fitness of a MEP chromosome is given in Table \ref{mep-proc-tab1}. Suppose that we have to solve a regression problem with a single example for training. Numerical values for inputs are as follows: $a = 7, b = 2, c = 1, d = 5$ and the expected output is 10.

\begin{table}
\caption{An example for the MEP fitness assignment process. The first column contains the expression encoded into the MEP chromosome. The second column contains the value of these expressions. These values are computed by replacing the variables with their values. The last column represents the fitness of each expression which is the difference in absolute value between value of each expression and the expected output for the current fitness case (which has the value 10 in our example).}
\label{mep-proc-tab1}
\begin{center}
\begin{tabular}
{p{70pt}p{70pt}p{50pt}}
\hline
Expressions& 
Value&
Fitness \\
\hline
$E_1 = a$&
7&
3\\
$E_2 = b$&
2&
8\\
$E_3 = a+b$&
9&
1\\
$E_4 = c$&
1&
9\\
$E_5 = d$&
5&
5\\
$E_6 = c+d$&
6&
4\\
$E_7 = (a+b)*d$&
45&
35\\
$E_8 = b*(c+d)$&
12&
2\\
\hline
\end{tabular}
\end{center}
\end{table}

Table \ref{mep-proc-tab1} provides a simple example on how to compute the fitness of a MEP chromosome. In this example the expression $E_3 =a+b$ provides the lowest value for fitness and thus this expression will represent the chromosome.

Let us take another example that contains two fitness cases. The first one is described by the values $a = 7, b = 2, c = 1, d = 5$ and the expected output is 10. The second fitness case is given by the inputs $a = 12, b = 3, c = 9, d = 1$ and the expected output is 7. Numerical values for this example are given in Table \ref{mep-proc-tab2}.

\begin{table}
\caption{Another example for the MEP fitness assignment process. The first column contains the expression encoded into the MEP chromosome. The second column contains the value of these expressions for the first fitness case. The third column contains the value of these expressions for the second fitness case. The last column represents the fitness of each expression which is the sum of the fitnesses for each of the two training cases.}
\label{mep-proc-tab2}
\begin{center}
\begin{tabular}
{p{70pt}p{70pt}p{70pt}p{70pt}}
\hline
Expressions& 
Value (case 1)&
Value (case 2)&
Fitness \\
\hline
$E_1 = a$&
7&
12&
3+5=8\\
$E_2 = b$&
2&
3&
8+4=12\\
$E_3 = a+b$&
9&
15&
1+8=9\\
$E_4 = c$&
1&
9&
9+2=11\\
$E_5 = d$&
5&
1&
5+6=11\\
$E_6 = c+d$&
6&
10&
4+3=7\\
$E_7 = (a+b)*d$&
45&
15&
35+8=43\\
$E_8 = b*(c+d)$&
12&
30&
2+23=25\\
\hline
\end{tabular}
\end{center}
\end{table}

The example shown in Table \ref{mep-proc-tab2} is a little bit more complicated. It contains two training (fitness) cases. For computing the fitness of an expression we had to compute the fitness for each training case and then to add these 2 values. The best expression (that will be used to represent the chromosome) is $E_6 = c+d$ in this example. It provided the lowest fitness among all expressions encoded in that chromosome as can be seen from Table \ref{mep-proc-tab2}.

In is obvious (see Tables \ref{mep-proc-tab1} and \ref{mep-proc-tab2} that some parts of a MEP chromosome are not used. Some GP techniques, like Linear GP, remove non-coding sequences of chromosome during the search process. As already noted \cite{brameier1} this strategy does not give the best results. The reason is that sometimes, a part of the 
useless genetic material has to be kept in the chromosome in order to 
maintain population diversity.\\

\section{MEP fitness in the case of multiple outputs problems}
\label{mep_fit_multiple_outputs}

In this section we describe the way in which Multi Expression Programming 
may be efficiently used for handling problems with multiple outputs (such as designing digital circuits \cite{oltean_circuits_nasa}).

Each problem has one or more inputs (denoted by \textit{NI}) and one or more outputs 
(denoted \textit{NO}). In section \ref{MEP_fitness} we have presented the way in which is the fitness of a 
chromosome with a single output is computed. 

When multiple outputs are 
required for a problem, we have to choose \textit{NO} genes which will provide the 
desired output (it is obvious that the genes must be distinct unless the 
outputs are redundant).

In Cartesian Genetic Programming (see chapter \textbf{CGP}), the genes providing the program's output are evolved just like all 
other genes. In MEP, the best genes in a chromosome are chosen to provide 
the program's outputs. When a single value is expected for output we simply 
choose the best gene (see section \ref{MEP_fitness}). 

When multiple 
genes are required as outputs we have to select those genes which minimize 
the difference between the obtained result and the expected output. 

We have to compute first the quality of a gene (sub-expression) for a given 
output (how good is that gene for providing the result for a given output):

\begin{equation}
\label{eq3}
f(E_i ,q) = \sum\limits_{k = 1}^n {\left| {o_{k,i} - w_{k,q} } \right|} ,
\end{equation}

\noindent
where $o_{k,i}$ is the obtained result by the expression (gene) $E_{i}$ for 
the fitness case $k$ and $w_{k,q}$ is the targeted result for the fitness case 
$k$ and for the output $q$. The values $f(E_{i}$, $q)$ are stored in a matrix (by using 
dynamic programming \cite{bellman1} for latter use (see formula (\ref{eq4})).

Since the fitness needs to be minimized, the quality of a MEP chromosome may be 
computed by using the formula:

\begin{equation}
\label{eq4}
f(C) = \mathop {\min }\limits_{i_1 ,i_2 ,...,i_{NO} } \sum\limits_{q = 1}^{NO} {f(E_{i_q } ,q)} .
\end{equation}

In equation (\ref{eq4}) we have to choose numbers $i_{1}$, $i_{2}$, \ldots , 
$i_{NO}$ in such way to minimize the program's output. For this we shall use 
a simple heuristic which does not increase the complexity of the MEP 
decoding process: for each output $q$ (1 $ \le q \le $ \textit{NO}) we choose the gene 
$i$ that minimize the quantity $f(E_{i}$, $q)$. Thus, to an output is assigned the 
best gene (which has not been assigned before to another output). The 
selected gene will provide the value of the $q^{th}$ output.\\

\textbf{Example}\\

Let's consider a problem with 3 outputs and a MEP chromosome with 5 genes. We compute the fitness of each gene for each output as described by equation \ref{eq3} and we store the results in a matrix. An example of matrix filled using equation \ref{eq3} is given in Table \ref{mep_multi_outputs_matrix}.

\begin{table}
\caption{An example on how to compute genes which will provides outputs. The problem has 3 outputs and the MEP chromosome has 5 genes. The value within cell ($i$, $j$) is the fitness provided by gene $i$ for output $j$ as computed by equation \ref{eq3}. Genes providing the outputs have been marked with *}
\label{mep_multi_outputs_matrix}
\begin{center}
\begin{tabular}
{|p{30pt}|p{40pt}|p{40pt}|p{40pt}|}
\hline
Gene\#& 
Output$_{1}$&
Output$_{2}$&
Output$_{3}$ \\
\hline
0&
5&
3&
9\\
\hline
1&
7&
6&
7\\
\hline
2&
1 $*$&
0&
2\\
\hline
3&
4&
1 $*$&
5\\
\hline
4&
2&
3&
4 $*$\\
\hline
\end{tabular}
\end{center}
\end{table}

The way in which genes providing the outputs have been selected is describes in what follows: first we need to compute the gene which will provide the result for the first output. We check the $2^{nd}$ column in Table \ref{mep_multi_outputs_matrix} and we see that the minimal value is  1 which correspond to the gene \#2. Thus the result for the first output of the problem will be provided by gene \#2. For the second output we see that gene \#2 is generating the lowest fitness. We cannot choose this gene again because it already provides the result for the first output. Thus we choose the gene \#3 for providing the result for the second output. Acording to the same algorithm (described above) the result for the third output will be provided by the gene \#4.\\

\textbf{Remark}\\

Formulas (\ref{eq3}) and (\ref{eq4}) are the generalization of formulas (\ref{eq1}) and (\ref{eq2}) for the case of multiple outputs of a MEP chromosome.

The complexity of the heuristic used for assigning outputs to genes is: \\

O(\textit{NG $ \cdot $ NO}), \\

\noindent
where \textit{NG} is the number of genes and \textit{NO} is the number of outputs.

We may use another procedure for selecting the genes that will provide the 
problem's outputs. This procedure selects, at each step, the minimal value 
in the matrix $f(E_{i}$, $q)$ and assign the corresponding gene $i$ to its paired 
output $q$. Again, the genes already used will be excluded from the search. 
This procedure will be repeated until all outputs have been assigned to a 
gene. However, we did not used this procedure because it has a higher 
complexity -- $O$(\textit{NO}$ \cdot $\textit{log}$_{2}$(\textit{NO})$ \cdot $\textit{NG}) - than the previously 
described procedure which has the complexity $O$(\textit{NO}$ \cdot $\textit{NG}). 

\section{MEP representation vs. GP, LGP, GEP and CGP representations}
\label{MEPvsOthers}

Generally a GP chromosome encodes a single expression (computer program). 
This is also the case for GEP, GE and LGP chromosomes. By contrast, a MEP 
chromosome encodes several expressions (as it allows a multi-expression 
representation). Each of the encoded expressions may be chosen to represent 
the chromosome, i.e. to provide the phenotypic transcription of the 
chromosome. Usually, the best expression that the chromosome encodes 
supplies its phenotypic transcription (represents the chromosome).

Although, the ability of storing multiple solutions in a single chromosome 
has been suggested by others authors, too (see for instance \cite{lones1}), and 
several attempts have been made for implementing this ability in GP 
technique. A related work could be considered the Handley's idea \cite{handley1} which stored the entire population of GP 
trees in a single graph. In this way a lot of memory is saved. 

\subsection{Multi Expression Programming versus Linear Genetic Programming}

There are many similar aspects of MEP and LGP representations. Both chromosomes can be interpreted as a sequence of instructions that can be translated into programs of an imperative language.

However, there are also several important aspects that makes these two methods different:

\begin{itemize}

\item{LGP modifies an array of variables. Each variable can be modified multiple times by different instructions. MEP computes the outputs generated by each gene and stores them into an array, but the obtained values are not overwritten by other values.}

\item{The variable providing the output in the case of LGP is chosen at the beginning of the search process, whereas, in the case of MEP, the gene providing the output could dynamically change at each generation.}

\end{itemize}

Linear GP \cite{brameier1} is also very suitable for storing multiple solutions in a 
single chromosome. In that case the multi expression ability is given by the 
possibility of choosing any variable as the program output. This idea is later exploited in  \textbf{Multi Solution Chapter}.

\subsection{Multi Expression Programming vs. Gene Expression Programming}

There are several major differences between GEP and MEP:

\begin{itemize}

\item{The pointers toward function arguments are encoded explicitly in the MEP chromosomes whereas in GEP these pointers are computed when the chromosome is parsed. MEP also needs to evolve the pointers toward function arguments. This leads to a more compact representation of GEP when compared to MEP.}

\item{MEP representation is suitable for code reuse, whereas GEP representation is not.}

\item{The noncoding regions of GEP are always at the end of the chromosome, whereas in MEP these regions can appear anywhere in the chromosome.}

\end{itemize}

It can be seen that the effective length of the expression may increases 
exponentially with the length of the MEP chromosome. This is happening because 
some sub-expressions may be used more than once to build a more complex (or 
a bigger) expression. Consider, for instance, that we want to obtain a 
chromosome that encodes the expression $a^{2^n}$, and only the operators 
{\{}+, -, *, /{\}} are allowed. If we use a GEP representation the 
chromosome has to contain at least (2$^{n + 1 }$-- 1) symbols since we need 
to store 2$^{n}$ terminal symbols and (2$^{n}$ -- 1) function operators. A 
GEP chromosome that encodes the expression $E=a^{8}$ is given below:\\

$C$ = *******\textit{aaaaaaaa}.\\

A MEP chromosome uses only (3$n$ + 1) symbols for encoding the 
expression $a^{2^n}$. A MEP chromosome that encodes expression $E = a^{8}$ is 
given below:\\

1: $a$

2: * 1, 1

3: * 2, 2

4: * 3, 3\\

As a further comparison, when $n$ = 20, a GEP chromosome has to have 2097151 
symbols, while MEP needs only 61 symbols. 

\subsection{Multi Expression Programming versus Cartesian Genetic Programming}

MEP representation is similar to GP and CGP, in the sense that each function 
symbol provides pointers towards its parameters. 

There are 3 main differences between MEP and CGP:

\begin{itemize}

\item{Within CGP the inputs are not stored in the chromosome. In MEP, the inputs are stored in any position of the chromosome. This fact is a consequence of the motivations that have lead to the proposal of these methods: CGP was proposed  for designing digital circuits whereas MEP was motivated by the way in which C and Pascal compilers translate mathematical expressions into machine code \cite{aho1}.}

\item{Within CGP the outputs are evolved along the other genes of the chromosome. In MEP all outputs are checked and the best of them is chosen to represent the chromosome.}

\item{CGP stores a single solution / chromosome, whereas MEP stores multiple solutions in the same chromosome. Note, that both procedures have the same complexity which means the same running time.}

\end{itemize}

\section{Search operators}
\label{MEP_Search_operators}

The search operators used within MEP algorithm are crossover and mutation. 
These search operators preserve the chromosome structure. All offspring are 
syntactically correct expressions. 

\subsection{Crossover}

By crossover two parents are selected and are recombined.

Three variants of recombination have been considered and tested within our 
MEP implementation: one-point recombination, two-point recombination and 
uniform recombination.\\

\subsubsection{One-point recombination}

One-point recombination operator in MEP representation is similar to the 
corresponding binary representation operator. One crossover point is 
randomly chosen and the parent chromosomes exchange the sequences at the 
right side of the crossover point.\\

\textbf{Example}\\

Consider the parents $C_{1}$ and $C_{2}$ given below. Choosing the crossover 
point after position 3 two offspring, $O_{1}$ and $O_{2}$ are obtained as 
given in Table \ref{mep_one_cut_point}.

\begin{table}[htbp]
\begin{center}
\caption{MEP one-point recombination.}
\begin{tabular}
{p{55pt}p{49pt}p{49pt}p{53pt}}
\hline
\multicolumn{2}{p{104pt}}{Parents } & 
\multicolumn{2}{p{103pt}}{Offspring}  \\
\hline
$C_{1}$& 
$C_{2}$& 
$O_{1}$& 
$O_{2}$ \\
\hline
1: \textbf{\textit{b}} \par 2: \textbf{* 1, 1} \par 3: \textbf{+ 2, 1} \par 4: \textbf{\textit{a}} \par 5: \textbf{* 3, 2} \par 6: \textbf{\textit{a}} \par 7: \textbf{- 1, 4}& 
1: $a$ \par 2: $b$ \par 3: + 1, 2 \par 4: $c$ \par 5: $d$ \par 6: + 4, 5 \par 7: * 3, 6& 
1: \textbf{\textit{b}} \par 2: \textbf{* 1, 1} \par 3: \textbf{+ 2, 1} \par 4: $c$ \par 5: $d$ \par 6: + 4, 5 \par 7: * 3, 6& 
1: $a$ \par 2: $b$ \par 3: + 1, 2 \par 4: \textbf{\textit{a}} \par 5: \textbf{* 3, 2} \par 6: \textbf{\textit{a}} \par 7: \textbf{- 1, 4} \\
\hline
\end{tabular}
\end{center}
\label{mep_one_cut_point}
\end{table}

\subsubsection{Two-point recombination}

Two crossover points are randomly chosen and the chromosomes exchange 
genetic material between the crossover points.\\

\textbf{Example}\\

Let us consider the parents $C_{1}$ and $C_{2}$ given below. Suppose that the 
crossover points were chosen after positions 2 and 5. In this case the 
offspring $O_{1}$ and $O_{2}$ are obtained as given in Table \ref{mep_two_cut_point}.

\begin{table}[htbp]
\begin{center}
\caption{MEP two-point recombination.}
\begin{tabular}
{p{55pt}p{49pt}p{49pt}p{49pt}}
\hline
\multicolumn{2}{p{104pt}}{Parents } & 
\multicolumn{2}{p{98pt}}{Offspring}  \\
\hline
$C_{1}$& 
$C_{2}$& 
$O_{1}$& 
$O_{2}$ \\
\hline
1: \textbf{\textit{b}} \par 2: \textbf{* 1, 1} \par 3: \textbf{+ 2, 1} \par 4: \textbf{\textit{a}} \par 5: \textbf{* 3, 2} \par 6: \textbf{\textit{a}} \par 7: \textbf{- 1, 4}& 
1: $a$ \par 2: $b$ \par 3: + 1, 2 \par 4: $c$ \par 5: $d$ \par 6: + 4, 5 \par 7: * 3, 6& 
1: \textbf{\textit{b}} \par 2: \textbf{* 1, 1} \par 3: + 1, 2 \par 4: $c$ \par 5: $d$ \par 6: \textbf{\textit{a}} \par 7: \textbf{- 1, 4}& 
1: $a$ \par 2: $b$ \par 3:\textbf{ + 2, 1} \par 4: \textbf{\textit{a}} \par 5: \textbf{* 3, 2} \par 6: + 4, 5 \par 7: * 3, 6 \\
\hline
\end{tabular}
\end{center}
\label{mep_two_cut_point}
\end{table}

\subsubsection{Uniform recombination}

During the process of uniform recombination, offspring genes are taken 
randomly from one parent or another.\\

\textbf{Example}\\

Let us consider the two parents $C_{1}$ and $C_{2}$ given below. The two 
offspring $O_{1}$ and $O_{2}$ are obtained by uniform recombination as 
given in Table \ref{mep_uniform}.

\begin{table}[htbp]
\caption{MEP uniform recombination.}
\label{mep_uniform}
\begin{center}
\begin{tabular}
{p{55pt}p{49pt}p{49pt}p{49pt}}
\hline
\multicolumn{2}{p{104pt}}{Parents } & 
\multicolumn{2}{p{98pt}}{Offspring}  \\
\hline
$C_{1}$& 
$C_{2}$& 
$O_{1}$& 
$O_{2}$ \\
\hline
1: \textbf{\textit{b}} \par 2: \textbf{* 1, 1} \par 3: \textbf{+ 2, 1} \par 4: \textbf{\textit{a}} \par 5: \textbf{* 3, 2} \par 6: \textbf{\textit{a}} \par 7: \textbf{- 1, 4}& 
1: $a$ \par 2: $b$ \par 3: + 1, 2 \par 4: $c$ \par 5: $d$ \par 6: + 4, 5 \par 7: * 3, 6& 
1: $a$ \par 2: \textbf{* 1, 1} \par 3: \textbf{+ 2, 1} \par 4: $c$ \par 5: \textbf{* 3, 2} \par 6: + 4, 5 \par 7: \textbf{- 1, 4}& 
1: \textbf{\textit{b}} \par 2: $b$ \par 3: + 1, 2 \par 4: \textbf{\textit{a}} \par 5: $d$ \par 6: \textbf{\textit{a}} \par 7: * 3, 6 \\
\hline
\end{tabular}
\end{center}
\end{table}

\subsection{Mutation}

Each symbol (terminal, function of function pointer) in the chromosome may 
be the target of the mutation operator. Some symbols in the chromosome are 
changed by mutation. To preserve the consistency of the chromosome, its 
first gene must encode a terminal symbol.

We may say that the crossover operator occurs between genes and the mutation 
operator occurs inside genes.

If the current gene encodes a terminal symbol, it may be changed into 
another terminal symbol or into a function symbol. In the later case, the 
positions indicating the function arguments are randomly generated. If the 
current gene encodes a function, the gene may be mutated into a terminal 
symbol or into another function (function symbol and pointers towards 
arguments).\\

\textbf{Example}\\

Consider the chromosome $C$ given below. If the boldfaced symbols are selected 
for mutation an offspring $O$ is obtained as given in Table \ref{mep_mutation}.

\begin{table}[htbp]
\caption{MEP mutation. Third and sixth genes have been affected by mutation}
\label{mep_mutation}
\begin{center}
\begin{tabular}
{p{70pt}p{70pt}}
\hline
$C$& 
$O$ \\
\hline
1: $a$ \par 2: * 1, 1 \par 3: \textbf{\textit{b}} \par 4: * 2, 2 \par 5: $b$ \par 6: +\textbf{ 3}, 5 \par 7: $a$& 
1: $a$ \par 2: * 1, 1 \par 3: \textbf{+ 1, 2} \par 4: * 2, 2 \par 5: $b$ \par 6: \textbf{+ 1}, 5 \par 7: $a$ \\
\hline
\end{tabular}
\end{center}
\end{table}

\section{Handling exceptions within MEP}\label{MEP_exceptions}

Exceptions are special situations that interrupt the normal flow of 
expression evaluation (program execution). An example of exception is 
\textit{division by zero,} which is raised when the divisor is equal to zero.

\textit{Exception handling} is a mechanism that performs special processing when an exception is 
thrown.

Usually, GP techniques use a \textit{protected exception} handling mechanism \cite{koza1}. For instance, if a 
division by zero exception is encountered, a predefined value (for instance 1 or the numerator) is returned.

MEP uses a new and specific mechanism for handling exceptions. When an 
exception is encountered (which is always generated by a gene containing a 
function symbol), the gene that generated the exception is mutated into a 
terminal symbol. Thus, no infertile individual may appear in a population.

There are some special issues that must be taken into account when implementing MEP for problems that might generate exceptions (such as division by 0). Because MEP chromosomes might be changed during the evaluation process it might be necessarily to recompute the some values for the previously (before the exception occurred) computed fitness cases.\\

\textbf{Example}\\

Let's consider a problem with 3 fitness cases and the current evaluated chromosome encodes the expression $a/b + b$. If the $3^{rd}$ fitness case is $a = 1$, $b = 0$, an exception will occur and the corresponding gene (which contains the function symbol /) will be mutated into a terminal symbol (either $a$ or $b$). In this case we need to recompute the value of the newly obtained expression for the fitness cases 1 and 2 and, of course, we have to compute its value for the remaining fitness case 3.

\begin{figure}[htbp]
\centerline{\includegraphics[width=3in,height=5.2in]{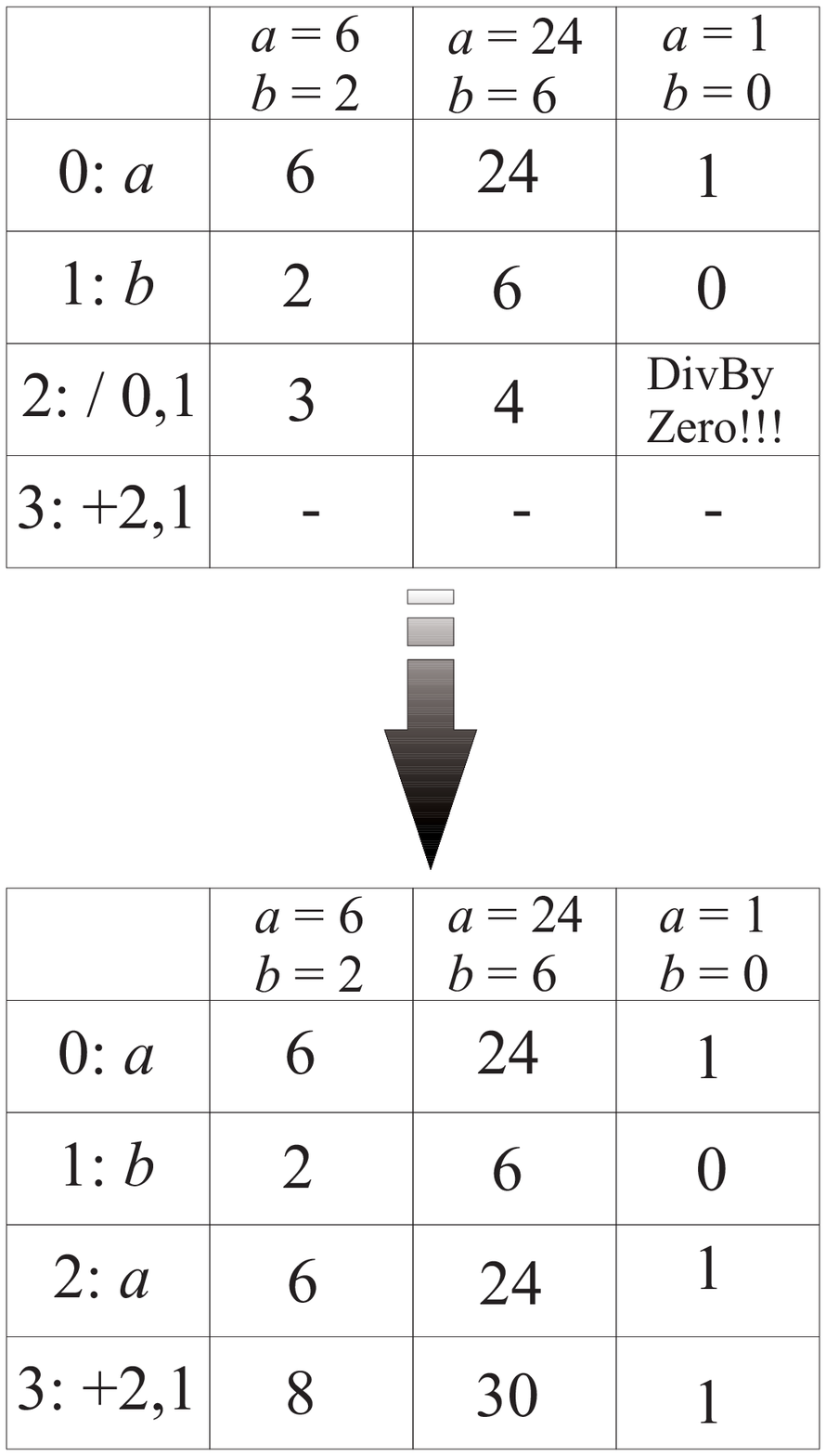}}
\label{mep_exceptions_fig}
\caption{An example on how to compute the value of the expressions encoded into a MEP chromosome. First column contains them MEP chromosome. First line contains three fitness cases. All other cells contain the value of the current expression for the current fitness case. An exception occurred in the $3^{rd}$ gene for the $3^{rd}$ fitness case (see table in the top of the figure). We need to recompute only the values for the gene which generated the exception ($4^{rd}$ row in our case). The values in all other rows (2 and 3) are preserved. At the end we may compute the values in the $4^{th}$ gene}
\end{figure}

For avoiding extra computations we may store partial results for all MEP genes and all fitness cases. Thus, we will need a $NumberOfGenes$ x $NumberOfFitnessCases$ matrix whose elements will be computed row by row. First of all we will compute the value of the expression encoded by the first gene for all fitness cases. Next, we move to the second gene and we compute the its value for all fitness cases. When an exception is encountered we need to recompute only the values of the current row. All other values (previously computed) remain unchanged. An example is depicted in Figure \ref{mep_exceptions_fig}.

The complexity of the storage space is:

\begin{center}
$O(NumberOfGenes * NumberOfFitnessCases)$.
\end{center}

\textbf{Remark}\\

For problems where exceptions are not encountered (e.g. digital circuits design) there is no need for storing this matrix. In this case we can compute the value of the expressions, encoded by a MEP chromosome, fitness case by fitness case. Only linear memory storage is required in this case.

\section{Complexity of the fitness computing process}
\label{MEP_complexity}

When solving symbolic regression, classification or any other problems for 
which the training set is known in advance (before computing the fitness), 
the fitness of an individual can be computed in O(\textit{CodeLength}) steps by dynamic 
programming \cite{bellman1}. In fact, a MEP chromosome needs to be read once for 
computing the fitness. 

Thus, MEP decoding process does not have a higher complexity than other GP - 
techniques that encode a single computer program in each chromosome. Roughly speaking this means the same running time for MEP and all other GP methods that parse the chromosome once in order to compute the fitness.

\section{MEP algorithm}
\label{MEP_algorithm}

Standard MEP algorithm uses steady-state evolutionary model \cite{syswerda1} as its underlying 
mechanism. 

The MEP algorithm starts by creating a random population of individuals. The 
following steps are repeated until a given number of generations is reached: 
Two parents are selected using a standard selection procedure. The parents 
are recombined in order to obtain two offspring. The offspring are 
considered for mutation. The best offspring $O$ replaces the worst individual 
$W$ in the current population if $O$ is better than $W$. 

The variation operators ensure that the chromosome length is a constant of 
the search process. The algorithm returns as its answer the best expression 
evolved along a fixed number of generations.

The standard MEP algorithm is outlined below:\\

\textbf{Standard MEP Algorithm}\\

\textsf{S}$_{1}$\textsf{. Randomly create the initial population 
}\textsf{\textit{P}}\textsf{(0)}

\textsf{S}$_{2}$\textsf{. }\textsf{\textbf{for}}\textsf{ 
}\textsf{\textit{t}}\textsf{ = 1 }\textsf{\textbf{to}}\textsf{ 
}\textsf{\textit{Max}}\textsf{{\_}}\textsf{\textit{Generations}}\textsf{ 
}\textsf{\textbf{do}}

\textsf{S}$_{3}$\textsf{. }\hspace{0.5cm}\textsf{\textbf{for}}\textsf{ 
}\textsf{\textit{k}}\textsf{ = 1 }\textsf{\textbf{to}}\textsf{ $\vert 
$}\textsf{\textit{P}}\textsf{(}\textsf{\textit{t}}\textsf{)$\vert $ / 2 
}\textsf{\textbf{do}}

\textsf{S}$_{4}$\textsf{. }\hspace{1cm}\textsf{\textit{p}}$_{1}$\textsf{ = 
}\textsf{\textit{Select}}\textsf{(}\textsf{\textit{P}}\textsf{(}\textsf{\textit{t}}\textsf{)); 
// select one individual from the current population}

\textsf{S}$_{5}$\textsf{. }\hspace{1cm}\textsf{\textit{p}}$_{2}$\textsf{ = 
}\textsf{\textit{Select}}\textsf{(}\textsf{\textit{P}}\textsf{(}\textsf{\textit{t}}\textsf{)); 
// select the second individual }

\textsf{S}$_{6}$\textsf{. }\hspace{1cm}\textsf{\textit{Crossover}}\textsf{ 
(}\textsf{\textit{p}}$_{1}$\textsf{, }\textsf{\textit{p}}$_{2}$\textsf{, 
}\textsf{\textit{o}}$_{1}$\textsf{, }\textsf{\textit{o}}$_{2}$\textsf{); // 
crossover the parents p}$_{1}$\textsf{ and p}$_{2}$

\hspace{4cm}\textsf{// the offspring o}$_{1}$\textsf{ and o}$_{2}$\textsf{ are obtained}

\textsf{S}$_{7}$\textsf{. }\hspace{1cm}\textsf{\textit{Mutation}}\textsf{ 
(}\textsf{\textit{o}}$_{1}$\textsf{); // mutate the offspring o}$_{1}$

\textsf{S}$_{8}$\textsf{. }\hspace{1cm}\textsf{\textit{Mutation}}\textsf{ 
(}\textsf{\textit{o}}$_{2}$\textsf{); // mutate the offspring o}$_{2}$

\textsf{S}$_{9}$\textsf{. }\hspace{1cm}\textsf{\textbf{if}}\textsf{ 
}\textsf{\textit{Fitness}}\textsf{(}\textsf{\textit{o}}$_{1}$\textsf{) $<$ 
Fitness(}\textsf{\textit{o}}$_{2}$\textsf{)}

\textsf{S}$_{10}$\textsf{. }\hspace{1cm}\textsf{\textbf{then}}\textsf{ 
}\textsf{\textbf{if}}\textsf{ 
}\textsf{\textit{Fitness}}\textsf{(}\textsf{\textit{o}}$_{1}$\textsf{) $<$ the 
fitness of the worst individual}

\hspace{5cm}\textsf{in the current population}

\textsf{S}$_{11}$\textsf{. }\hspace{1cm}\textsf{\textbf{then}}\textsf{ 
}\textsf{\textit{Replace}}\textsf{ the worst individual with 
}\textsf{\textit{o}}$_{1}$\textsf{;}

\textsf{S}$_{12}$\textsf{. }\hspace{1cm}\textsf{\textbf{else}}\textsf{ 
}\textsf{\textbf{if}}\textsf{ 
}\textsf{\textit{Fitness}}\textsf{(}\textsf{\textit{o}}$_{2}$\textsf{) $<$ the 
fitness of the worst individual}

\hspace{5cm}\textsf{in the current population}

\textsf{S}$_{13}$\textsf{. }\hspace{1cm}\textsf{\textbf{then}}\textsf{ 
}\textsf{\textit{Replace}}\textsf{ the worst individual with 
}\textsf{\textit{o}}$_{2}$\textsf{;}

\textsf{S}$_{14}$\textsf{. }\hspace{1cm}\textsf{\textbf{endfor}}

\textsf{S}$_{15}$\textsf{. }\textsf{\textbf{endfor }}

\section{Automatically Defined Functions in MEP}
\label{mep_adfs}

In this section we describe the way in which the Automatically Defined 
Functions \cite{koza2} are implemented within the context of Multi Expression 
Programming.

The necessity of using reusable subroutines is a day-by-day demand of the 
software industry. Writing reusable subroutines proved to reduce:

\begin{itemize}

\item[{\it (i)}]{the size of the programs.}

\item[{\it (ii)}]{the number of errors in the source code.}

\item[{\it (iii)}]{the cost associated with the maintenance of the existing software.}

\item[{\it (iv)}]{the cost and the time spent for upgrading the existing software.}

\end{itemize}

As noted by Koza \cite{koza2} function definitions exploit the underlying 
regularities and symmetries of a problem by obviating the need to tediously 
rewrite lines of essentially similar code. Also, the process of defining and 
calling a function, in effect, decomposes the problem into a hierarchy of 
subproblems.

A function definition is especially efficient when it is repeatedly called 
with different instantiations of its arguments. GP with ADFs have shown 
significant improvements over the standard GP for most of the considered 
test problems \cite{koza1,koza2}.

An ADF in MEP has the same structure as a MEP chromosome (i.e. a string of 
genes). The ADF is also evolved in the same way as a standard MEP chromosome. The function symbols used by an ADF are the same as those used by the standard MEP chromosomes. The terminal symbols used by an 
ADF are restricted to the function (ADF) parameters (formal parameters). For 
instance, if we define an ADF with two formal parameters $p_{0}$ and 
$p_{1}$ we may use only these two parameters as terminal symbols within the 
ADF structure, even if in the standard MEP chromosome (i.e. the main 
evolvable structure) we may use, let say, 20 terminal symbols only.\\

The set of function symbols of the main MEP structure is enriched with the 
Automatically Defined Functions considered in the system.\\

\textbf{Example}\\

Let us suppose that we want to evolve a program using 2 ADFs, denoted $ADF_0$ and $ADF_1$ having 2 ($p_{0}$ and $p_{1}$) respectively 3 ($p_{0}$ and $p_{1}$ and $p_{2}$) arguments. Let us also suppose that the terminal set for the main MEP chromosome is  $T$ = {\{}$a$, $b${\}} and the function set $F$ = {\{}+, -, *, /{\}}. The terminal and function symbols that may appear in ADFs and main MEP chromosome are given in Table \ref{table:16-5}.

\begin{table}[ht]
\begin{center}
\caption{Parameters, terminal set and the function set for the ADFs and for the main MEP chromosome}
\label{table:16-5}
\begin{tabular}
{p{75pt}p{53pt}p{70pt}p{115pt}}
\hline
& 
\textbf{Parameters}& 
\textbf{Terminal set}& 
\textbf{Function set} \\
\hline
$ADF_0$& 
$p_{0}$, $p_{1}$& 
$T$={\{}$ p_{0}$, $p_{1}${\}}& 
$F$={\{}+,--,*,/{\}} \\
$ADF_1$& 
$p_{0}$, $p_{1}$, $p_{2}$& 
$T$={\{}$p_{0}$, $p_{1}$, $p_{2}${\}}& 
$F$={\{}+,--,*,/{\}} \\
MEP chromosome& 
--& 
$T$={\{}$a$, $b${\}}& 
$F$={\{}+,--,*,/, $ADF_0$, $ADF_1${\}} \\
\hline
\end{tabular}
\end{center}
\end{table}

The $ADF_0$ could be defined as follows:\\

$ADF_0$ ($p_{0}$, $p_{1}$)

1: $p_{0}$

2: + 1, 1

3: $p_{1}$

4: / 3, 2

5: * 2, 4\\

The main MEP chromosome could be the following:\\

1: $a$

2: $b$

3: + 1, 2

4: $ADF_0$ 3, 1

5: $a$

6: $ADF_1$ 4, 5, 5

7: * 3, 6\\

The fitness of a MEP chromosome is computed as described in section \ref{MEP_fitness}.

The quality of an ADF is computed in a similar manner. The ADF is read once 
and the partial results are stored in an array (by the means of Dynamic 
Programming \cite{bellman1}). The best expression encoded in the ADF is chosen to 
represent the ADF. 

The genetic operators (crossover and mutation) used in conjunction with the 
standard MEP chromosomes may be used for the ADFs too. The probabilities for 
applying genetic operators are the same for MEP chromosomes and for the 
Automatically Defined Functions. The crossover operator may be applied only 
between structures of the same type (that is ADFs having the same parameters or main MEP chromosomes) in order to preserve the chromosome consistency.

\section{Applications}

Multi Expression Programming has been applied, so far, to the following problems:

\begin{itemize}

\item{Symbolic regression, classification \cite{oltean_complex}}
\item{Data analysis for real world problems \cite{oltean_complex}}
\item{Digital circuits design (adders, multipliers, even-parity) \cite{oltean_parity_fea,oltean_circuits_nasa,oltean_parity_trail}}
\item{Designing digital circuits for NP-Complete problems \cite{oltean_knapsack}}
\item{Designing reversible digital circuits \cite{oltean_reversible}}
\item{Evolving play strategies for Nim game \cite{oltean_nim}}
\item{Evolving evolutionary algorithms \cite{oltean_mep_eea,oltean_evolving_patterns}}
\item{Evolving heuristics for TSP problems \cite{oltean_tsp}}
\item{Modeling of Electronic Hardware \cite{ajith5}}

\end{itemize}

\section{Online resources}

More information about MEP can be found in the following web pages:

\begin{itemize}

\item{Multi Expression Programming, \url{https://mepx.org} or \url{https://mepx.github.io}}

\end{itemize}

\section{Tips for implementation}
\label{MEP_tips}

As MEP encodes multiple solutions within a chromosome it is important to set an optimal value for the chromosome length. The chromosome length directly affects the size of the search space and the number of solutions explored within this search space. Numerical experiments described in some of the next chapters show that it is a good idea to maintain chromosomes longer than the size of the actual solution for the problem being solved. If we set the chromosome length equal to the length of the (shortest) solution we will get a very low success rate. The success of MEP will increase (up to a point) as long as the length of the chromosome will be increased. However, the speed of the algorithm will decrease by increasing the length of the chromosome. Thus an optimal length of the chromosome must be a tradeoff between running time and the quality of the solutions.

\section{Summary}
\label{mep_summary}

Multi Expression Programming technique has been described in this chapter. 

A distinct feature of MEP is its ability to encode multiple solutions in the same chromosome. It has been shown that the complexity of the decoding process is the same as in the case of other GP techniques encoding a single solution in the same chromosome. Several numerical examples have been presented for the MEP decoding process.

A set of genetic operators (crossover and mutation) has been described in the context of Multi Expression Programming.

\section*{Problems}
\label{MEP_problems}
\addcontentsline{toc}{section}{Problems} 

\begin{itemize}

\item{Improve the speed of MEP by using Sub Machine Code GP \cite{poli1,poli2}.}

\item{Propose new techniques for reducing the duplicate expressions in MEP. This will reduce the size of the search space and will help MEP to perform better when solving difficult problems.}

\item{Implement and compare different mechanisms for handling exceptions in MEP. The current one (described in section \ref{MEP_exceptions}) mutates the gene that has generated the exception. Compare this mechanism with that used by LGP \cite{brameier1} or standard GP \cite{koza1}. Compare the running time and the evolution of the number of exceptions in these cases.
}

\item{Implement a MEP variant in which the chromosomes have variable length. Also propose some genetic operators that can deal with variable length MEP chromosomes. In this case the algorithm must be able to evolve the optimal length for the MEP chromosome. There are some difficulties related to this approach since MEP chromosomes will tend to increase their length in order to accommodate more and more solutions. And this could lead to bloat }

\item{In our implementation all symbols (terminals and functions) have the same probability to appear into a chromosome. There could be some problems in this case. For instance, if our problem has many variables (let's say 100) and the function set has only 4 symbols we cannot get too complex trees because the function symbols have a reduced chance to appear in the chromosome. Change the implementation (the procedures for initialization and mutation) so that the set of terminals and the set of functions have the same probability of being chosen in the chromosome. In this case one must choose first the set (either terminals or functions) and then one will randomly choose a symbol from that set. Compare those two implementation for different test problems.
}

\item{A variant of MEP is to keep all terminals in the first positions. No other genes containing terminals are allowed in the chromosome. Try to compare the standard variant with this one for some benchmark problems.
}

\end{itemize}

\end{document}